%% file: mcmcCnn.tex
\documentclass[times,twocolumn,final]{elsarticle}

\input{preamble_gth818n}

\usepackage{medima}
\usepackage{framed,multirow}
\usepackage{booktabs}

\usepackage{xcolor}
\usepackage{hyperref}
\usepackage{latexsym}

\definecolor{newcolor}{rgb}{.8,.349,.1}

\journal{in submission}

\begin{document}

\verso{Given-name Surname \textit{et~al.}}

\begin{frontmatter}

\title{MCMC Guided CNN Training and Segmentation for Pancreas Extraction}

\author[a]{Jinchan \snm{He}\fnref{fn1}}
\author[a]{Xiaxia \snm{Yu}\fnref{fn1}}
\fntext[fn1]{J. He and X. Yu contribute equally to this work.}
\author[b]{Chudong \snm{Cai}}
\author[a]{Yi \snm{Gao}\corref{cor1}}
\cortext[cor1]{Send corresponds to:}
\ead{gaoyi@szu.edu.cn}

\address[a]{School of Biomedical Engineering, Health Science Center, Shenzhen University, Shenzhen, Guangdong, China}
\address[b]{Department of General Surgery, Shantou Central Hospital and The Affiliated Shantou Hospital of Sun Yat-sen University, Shantou, Guangdong, China}

\received{1 May 2013}
\finalform{10 May 2013}
\accepted{13 May 2013}
\availableonline{15 May 2013}

%%\author{Jinchan~He\textsuperscript{*}, Xiaxia~Yu\textsuperscript{*}, Chudong~Cai, and Yi~Gao\textsuperscript{+}% <-this % stops a space
%%  \thanks{J. He, X. Yu, and Y. Gao are with the School of Biomedical Engineering, Health Science Center, Shenzhen University, Shenzhen, China.}
%%  \thanks{C. Cai is with the Department of General Surgery, Shantou Central Hospital and The Affiliated Shantou Hospital of Sun Yat-sen University, Shantou, Guangdong, China.}
%%  \thanks{\textsuperscript{*}J. He and X. Yu contribute equally to this work. \textsuperscript{+}Send corresponds to Y. Gao~(gaoyi@szu.edu.cn)}}% <-this % stops an unwanted space

\begin{abstract}
    Effificient organ segmentation is the precondition of various
    quantitative analysis. Segmenting the pancreas from abdominal CT
    images is a challenging task because of its high anatomical
    variability in shape, size and location. What's more, the pancreas
    only occupies a small portion in abdomen, and the organ border is
    very fuzzy. All these factors make the segmentation methods of
    other organs less suitable for the pancreas segmentation. In this
    report, we propose a Markov Chain Monte Carlo~(MCMC) sampling
    guided convolutionation neural network~(CNN) approach, in order to
    handel such difficulties in morphologic and photometric
    variabilities. Specifically, the proposed method mainly contains
    three steps: First, registration is carried out to mitigate the
    body weight and location variability. Then, an MCMC sampling is
    employed to guide the sampling of 3D patches, which are fed to the
    CNN for training. At the same time, the pancreas distribution is
    also learned for the subsequent segmetnation. Third, sampled from
    the learned distribution, an MCMC process guides the segmentation
    process. Lastly, the patches based segmentation is fused using a
    Bayesian voting scheme. This method is evaluated on the NIH
    pancreatic datasets which contains 82 abdominal contrast-enhanced
    CT volumes. Finally, we achieved a competitive result of 78.13\%
    Dice Similarity Coeffificient value and 82.65\% Recall value in
    testing data.
\end{abstract}


\end{frontmatter}

%%%%%%%%%%%%%%%%%%%%%%%%%%%%%%%%%%%%%%%%%%%%%%%%%%%%%%%%%%%%%%%%%%%%%%
%%%%%%%%%%%%%%%%%%%%%%%%%%%%%%%%%%%%%%%%%%%%%%%%%%%%%%%%%%%%%%%%%%%%%%

\section{Introduction}\label{sec:introduction}
Accurate organ segmentation is the prerequisite of many subsequent
computer based analysis. In recent years, with the rapid development
of deep neural network, automatic segmentation of many organs and
tissues have achieved good results, such as segmentation of the
cortical and sub-cortical structures, lung, liver, heart, 
etc..~\cite{lungSeg2018,lunglobeSeg2019,liverSeg2018,livertumorSeg2018,heartSeg2016,heartSeg2017}.
For the pancreas, however, while the segmentation result has been improved
substantially from the pre-deep learning era, the accuracy is still
relatively low compared with other organs~\cite{fixedPoint2017,hierarchical2018,deeporgan2015,bottomUp2017,spatial2015,spatialLocation2018}. 
This is mainly caused by several aspects: 1) the shape, size and 
position of the pancreas vary greatly among different patients in 
abdomen; 2) the contrast between the pancreas and surrounding 
tissue is weak; 3) pancreas is relatively soft and easy to be 
pushed by surrounding organs, which could result in large 
deformation; 4)pancreas occupies a very small portion in the
CT image. All these factors make accurate segmentation still a
challenging task.

%%%%%%%%%%%%%%%%%%%%%%%%%%%%%%%%%%%%%%%%%%%%%%%%%%%%%%%%%%%%%%%%%%%%%%
\subsection{Related works}\label{sec:introduction:relatedwork}
In recent years, more attention has been paid to to the segmentation
of pancreas.  Many pancreas segmentation methods have been proposed.
In~\cite{multiAtlas2017}, authors adopted a multi-atlas framework for
the pancreas segmentation. Specifically, the region of pancreas is
extracted by the relative position and structure of pancreas and
liver. Then, using the vessel structure around the pancreas, images
from a training dataset are registrated to image to be
segmented. Then, the best registration, evaluated based on the
similarity of the vessel structure, is chosen. This method reports a
result with a dice similarity coefficient~(DSC) of
78.5$\pm$14.0\%.

%% But this kind of methods which are based on
%% multi-atlas depends on bigger datasets, and the deviation of this
%% method is high.

Due to the low constrast nature of the pancreas, more advanced deep
learning and machine learning approaches are used. Such as using CNN 
in classification~\cite{fixedPoint2017,hierarchical2018} 
as well as a combination of CNN and random 
forest~\cite{deeporgan2015,bottomUp2017,spatial2015,spatialLocation2018}.

Unlike the top-down approach based on multi-atlas
in~\cite{multiAtlas2017}, researchers
in~\cite{bottomUp2014,bottomUp2017} proposed a bottom-up pancreas
segmentation strategy. It decomposes all 2D slices of a patient into
boundary-preserving superpixels by over segmentation. Then, it
extracts superpixel-level feature from the original CT image slices
and the dense image patch response maps to classify superpixels as
pancreas and non-pancreas by using a a two-level cascade random forest
classifier. Comparing with~\cite{multiAtlas2017}, these methods have
less data requirements, but the results have a slightly lower DSC of
68.8$\pm$25.6\% in~\cite{bottomUp2014} and 70.7$\pm$13.0\%
in~\cite{bottomUp2017}.

Similar to~\cite{bottomUp2017}, the methods proposed by
~\cite{deeporgan2015,spatial2015,spatialLocation2018} all combine the
random forest classification and the CNN. In~\cite{deeporgan2015},
authors presents a coarse-to-fine approach in which multi-level CNN is
employed on both image patches and regions. In this approach, an
initial set of superpixel regions are generated from the input CT
images by a coarse cascade process of random forests based
on~\cite{bottomUp2014}. Serving as regional candidates, these
superpixel regions possess high sensitivity but low precision. Next,
trained CNN are used to classify superpixel regions as pancreas and
non-pancreas. 3D Gaussian smoothing and 2D conditional random fields
are used for post-processing finally. Different
from~\cite{deeporgan2015}, researchers in~\cite{spatial2015} proposed
a method which is using random forest classification to classify
superpixels. This is similar to~\cite{bottomUp2014}
and~\cite{bottomUp2017}. But the superpixel and feature are generated
via Holistically-Nested Networks, which extract the pancreas’ interior
and boundary mid-level cues.  Before this step, this model gets the
interesting region by the method in~\cite{bottomUp2014}. Based
on~\cite{spatial2015}, authors in~\cite{spatialLocation2018} made some
improvements, and the major improvment is that that they get the 
interesting region by a new general deep learning based approach. 
In~\cite{spatialLocation2018}, the algorithm learns mid-level cues 
via Holistically-Nested Networks firstly. Then, it obtains the 
interesting region by a multi-view aggregated Holistically-Nested 
Networks and the largest connected component analysis. Finally, the 
random forest classification is adopted to classify superpixels, too.

There have also been some methods purely using the CNN for the
segmentation. For example, a fixed-point model that shrinks the input
region by the predicted segmentation mask was proposed
in~\cite{fixedPoint2017}. While the parameters of network remain
unchanged, the regions are optimized by an iterative process.
Contrasting to~\cite{fixedPoint2017}, the approach
in~\cite{hierarchical2018} pays more attention to the structure of
network: It proposed a novel network which is based on the Richer
Feature Convolutional Network. This network replaces the simple
up-sampling operation into multi-layer up-sampling structure in all
stages.

%%%%%%%%%%%%%%%%%%%%%%%%%%%%%%%%%%%%%%%%%%%%%%%%%%%%%%%%%%%%%%%%%%%%%%
\subsection{Contribution of this paper}\label{sec:introduction:contribution}
In this paper, we propose a robust and efficient segmentation approach
based the Markov Chain Monte Carlo~(MCMC) guided CNN. The proposed
aproach mainly consists of three parts: First, it locates pancreas in
the abdominal CT image by image registraion, we get a irregular
candidate region with a large number of background pixels being
rejected and almost all foreground pixels being preserved; Then, the
MCMC samples guide the training of the 3D CNN to classify pixels in
the candidate region; Finally in the segmentation, the MCMC guides the
trained network to perform the fused segmentation across all the image
domain.

The contributions of our work are mainly summarized in the following four
points: 1) Inspired by coarse to fine segmentation method and considered
the fact that pancreas' region is small in abdomen, we try to use
registration method to get a coarse segmentation to locate pancreas
in abdomen, but in order to improve the accuracy of this coarse
segmentation, we use the method of multi-atlas; 2) Adding some background
pixels to the coarse segmentation's pancreas region, we generate a irregular
candidate region which is like a bounding box, and the number of background
pixels and foreground pixels which are entered into the CNN data will
be more balanced because of these process; 3) Comparing with 2D CNN, 3D
CNN can use the information between slices, but 3D CNN faces the problem
of insufficient data. We fetch 3D image patches from candidate region to
training CNN instead of the whole candidate region to avoid this
problem ; 4) When training and testing CNN, we employ Markov Chain Monte
Carlo method to guide this precess focusing more on the irregular candidate
region so that CNN can focus on learn and test the internal and marginal
features of pancreas shielding the interference of surrounding organs
and tissues.

The remaining parts of the manuscript are organized as follows: the
proposed learning and segmetnation framework is detailed in
Section~\ref{sec:method}. Then, experiments are conducted on 
data set in Section~\ref{sec:some}. Finally, the work is 
concluded with furture direction discussed in Section~\ref{sec:discussion}.

%%The remaining parts of the manuscript are organized as follows: the
%%proposed learning and segmetnation framework is detailed in
%%Section~\ref{sec:method}. Then, experiments are conducted on several
%%data sets\todo{to be determiend later} in
%%Section~\ref{sec:some}. Finally, the work is concluded with furture
%%direction discussed in Section~\ref{sec:discussion}.

%%%%%%%%%%%%%%%%%%%%%%%%%%%%%%%%%%%%%%%%%%%%%%%%%%%%%%%%%%%%%%%%%%%%%%%%%%%%%%%%
\section{Method}\label{sec:method}
In this section, the proposed MCMC guided CNN learning and
segmentation framework is detailed. Specifically, a Markov Chain Monte
Carlo process is utilized to guide the learning in the sample space
focusing more on the target region in
Section~\ref{sec:method:learn}. After the learning, the segmentation
is also governed by the optimal filtering, which results in a robust
and fast 3D segmentation, detailed in Section~\ref{sec:method:seg}.

%%%%%%%%%%%%%%%%%%%%%%%%%%%%%%%%%%%%%%%%%%%%%%%%%%%%%%%%%%%%%%%%%%%%%%
\subsection{Joint learning of appearence and location}\label{sec:method:learn}
Denote the image to be segmented as $I:\bR^3 \to \bR$. The
segmentation of $I$ is seeking for a indicator function $J:\bR^3\to
\{0,1\}$ whose 0 valued pixels indicate the background and the 1s
indicate the pancreas. Such a characteristic function $J$ can be viewed
as a special case of the probability density function~(pdf)
$p:\bR^3\to [0,1]$ where the value is a measure for the pixel being
inside the hippocampus.

Viewed this way, the pdf $p$ can be estimated using a filtering based
convolutional neural network framework. Specifically, the samples are
drawn from $p$ based on an MCMC process, which guides the CNN to train
on the key regions of the target as well as the border area. Then in
the segmentation stage, the same MCMC process again guide the trained
CNN to sample and segmentation each patches. This effectively avoid
sampling from the entire image domain. Instead, it only learn and
discriminate on the regions with higher probability of finding the
target. The iterative process provides a final accuracy segmentation
for the pancreas in $I$ by fusing the patch segmentations.

More explicitly, given $\bx\in \bR^3$ and its neighborhood
$N(\bx):=\{\by\in\bR:\|\bx - \by\|_{\infty} \le h\}$ of radius $h$, we
restrict $I(\bx)$ on $N(\bx)$ to define $L(\bx):N(\bx)\to \bR$ with
$L(\bx):=I(\bx)|_{N(\bx)}$. Then, the probability $p(L(\bx), \bx)$
ultimately provides the information whether the point $\bx$ belongs to
the pancreas in $I$.

Apparently, we have
\begin{align}
  p(L(\bx), \bx) = p(L(\bx)| \bx)\cdot p(\bx)
  \label{eq:bayes}
\end{align}
where $p(\bx)$ is the prior distribution and the $p(L(\bx)| \bx)$
depicts the probability of a certain patch $L(\bx)$ being inside
the pancreas, conditioned on its spatial location. The goal is to
obtain an estimation for $p(L(\bx), \bx)$.

To that end, we have a set of training images $I_i:\bR^3\to \bR, i=1,
\dots, M$ segmented with their pancreas segmentation denoted as
$J_i:\bR^3\to \{0,1\}, i=1, \dots, M$. The training set provides us
with the location and contexual information for the
pancreas. Specifically, the prior distribution $p(\bx)$ in
Equation~\ref{eq:bayes} can be learned from $J_i$ directly. Moreover,
the second term will be learned through an optimal filter guided
CNN.

To proceed, it is realized that the training set has a large variance
on the shape and size of the pancreas. Ideally, all such variations
could be learned with the above framework. However,
normalization registration would be helpful to reduce the variance and
leaves less burdern on the learning.

Before the emerging of the convolutional neural network approach,
multi-atlas is a robust algorithm that addresses the problem of
medical image segmentation by (multiple) registrations~\cite{RN487}. It
achieved very high segmentation accuracy, especially for the brain
structures~\cite{RN466,RN276,corticalAtlasBrain2016,harmonizedAtlasBrain2018,supervoxelAtlasBrain2018}.
Unfortunately, comparing to brain segmentation, the performance
on other sites is much worse.

Such discrepency is understandable: the shape, size, and image
appearence in abdominal images vary much severely than that in the
brain image. As a result, non-linear registration performs worse on
abdominal images.

Based on this rationale, in this study we only use affine
registration among the images to mitigate the training
variance, leaving the rest to the machine learning framework.

To proceed, a random training image is picked from $I_i:i=1, \dots, M$
and we denote the image as $\tilde{I}$ and its segmentation as
$\tilde{J}$. Then, we have the registration computated as:
\begin{align}
  T^i = \argmin_{T:\bR^3\to\bR^3} D(\tilde{I}, I_i\circ T)
\end{align}
where $T$ can be written as $T(\bx) = A\bx + b$ with
$A\in\bR^{3\times 3},b\in\bR^3$. $D(I_i, I_j)$ denotes suitable
dis-similarity functional between the two images $I_i$ and $I_j$.

The optimal registration transformations $T^i$ can be computed through
regular gradient or Newton based procedures. As a result,
$\tilde{I}\approx I_i\circ T^i =:\tilde{I}_i, \; \forall i$ and
$\tilde{J}\approx J_i\circ T^i =:\tilde{J}_i, \; \forall i$.

Once registered to a common space, the collection of $\tilde{J}_i$
represent the spatial distribution of the pancreas in the training
data. Specifically, compute $P:\bR^3\to \bR^+$ as :
\begin{align}
  \label{eq:spatialPrior}
  P := \left. \sum_{i=1}^M \tilde{J}_i \middle/ \int \sum_{i=1}^M \tilde{J}_i d\bx \right.
\end{align}
and we get the prior distribution of the pancreas in $P$.

\subsection{MCMC sampling guided training with 3D CNN}\label{sec:method:train}
Learning the segmentations of patches from 2D or 3D images and applying
the learned model patch-wise for the testing image, is a common
approaches in learning-based segmentation.

However, one difficulty is how the samples are drawn from the entire
image domain. If the patches are taken uniformly from the entire image
domain, then a large portion of the patches are empty (all-zero
mask). The learning of such patches is therefore a process of mapping
certain grayscale image to an all-zero image using, for example,
convolutional neural network.

Unfortunately, learning such empty masks does more harms than
wasting time. Indeed, it has a trivial global optimal solution: When
all the parameters are set to zero, the output will be an empty
mask. Especially when the area of the object takes a small portion of
the entire image domain, uniformly aquiring patches will result in
such a situation. Therefore, many researchers use an extra mechanism,
such as a bounding box, to limit the volume from where the patches are
generated. However, how to determine such a bounding box for the
testing image also poses new problem.

Therefore, on one hand, we want the training patches to contain
a certain amount of empty patches to learn the negative appearence. On
the other hand, we don't want too many of them to stear the learn
towards an unwanted global optimal.

In order to address such an issue, we propose to use the prior
distribution $P(\bx)$ defined in Eq~\ref{eq:spatialPrior} to guide the
sampling process. Indeed, the high value in $P(\bx)$ is an indication
of being more likely to be inside the pancreas. Consequently, the
image appearence in those regions are more typical for
pancreas. Therefore, in the training process, more emphasis should be
put on those regions. Specifically, one can draw samples
$\{\bm{s}_i\in \bR^3:\bm{s}_i \sim P(\bx) \}$. Then, patches can be
taken around $\bm{s}_i$'s and be fed to the neural network for
learning.

Due to the arbitrariness of $P(\bx)$, one can't use regular sampling
schemes as for a normal distribution. In this work, we use the Markov
Chain Monte Carlo draw samples from $P(\bx)$.

Markov chain Monte Carlo (MCMC) is an class of algorithms, which
are often applied to produce samples from multi-dimensional
probability distribution, and these samples can approximate this
original probability distribution~\cite{mcmc1970}. We construct a Markov chain
based on a stationary probability distribution, which is equal
to the original probability distribution firstly when we
draw samples by MCMC. Then we draw samples by this Markov
chain. Finally we can get the target samples when the Markov chain
achieves a stationary distribution.

Specifically, the Metropolis-Hastings~(MH) algorithm is used in this
work to obtain random samples from $P(\bx)$ which is hard to sample
directly. The main steps of MH guided patch generation
algorithm~\cite{mcmcMH1995} are detailed as follows.
\begin{algorithm}
  \caption{Metropolis-Hastings guided patch generation}
  \begin{algorithmic}[1]
    \STATE{Input a state transition matrix $q$ of this markov chain and stationary probability distribution $P(\bx)$}
    \STATE{Set the threshold of the times of state transition as $n_1$ and the needed number of samples as $n_2$}
    \STATE{Generate an initial state $\bx_0$}
    \FOR{$t=0, 1, 2, ..., n_1+n_2-1$}
    \STATE{Generate a proposal state $\bx^*$ from $q(\bx|\bx_t)$}
    \STATE{Draw a random number $u$ from $Uniform(0,1)$}
    \STATE{Calculate the acceptance probability
        \begin{align}
          \label{eq:acceptprob}
          \alpha(\bx_t,\bx^*) = min\{\frac{p(\bx^*)q(\bx^*,\bx_t)}{p(\bx_t)q(\bx_t,\bx^*)},1\}\nonumber
        \end{align}
    }
    \IF{$u < \alpha(\bx_t,\bx^*)$}
    \STATE{accept the proposal, set $\bx_{t+1} = \bx^*$, extract a 3D patch at $\bx_{t+1}$}
    \ELSE
    \STATE{$\bx_{t+1} = \bx_t$}
    \ENDIF
    \ENDFOR
  \end{algorithmic}
  \label{algo:mcmc-mh}
\end{algorithm}

After some steps $n_1$, the samples
$\{\bx_{n_1},\bx_{n_1+1},\dots,\bx_{n_1+n_2-1}\}$ correspond to the
stationary probability distribution $P(\bx)$. Once the samples
$\bm{s}_i$'s are drawn, 3D patches $Q_i$ of the size $z$ are taken
around each $\bm{s}_i$:
\begin{align}
  Q_i(\bx) = \left. I_j(\bx)\middle|_{\bx \in N(\bm{s}_i)}\right. \nn
  \text{where } j \sim U(\{1, \dots, M\}).
\end{align}
These 3D patches are then set as the input for the 3D CNN to learn.

Various convolutional neural network architectures have been proposed
in the recent years~\cite{alexnet2012,vggnet2014,googlenet2014,resnet2015,unet2015,3dunet2016}. Amoung them, the
U-Net is a CNN framework for segmentation~\cite{unet2015} which have
achieved substantial success in particular the biomedical image
segmentation field. U-Net adopts the symmetric encoder-decoder
structure. In the process of encoder, the image was down-sampled
several times, the size of feature map becomes smaller, but the
feature channels increase. After this encoding, the network captures
the low-level information of the image. In the process of decoder, the
image was up-sampled four times. After this up-sampling, the network
captures the high-level information of the image. Both the low-level
information and the high-level information of the image are captured.
This symmetrical structure has achieved good performance in the
applications of various biomedical segmentation.

\begin{figure*}[htb]
  \centering
  \includegraphics[width=6.4in]{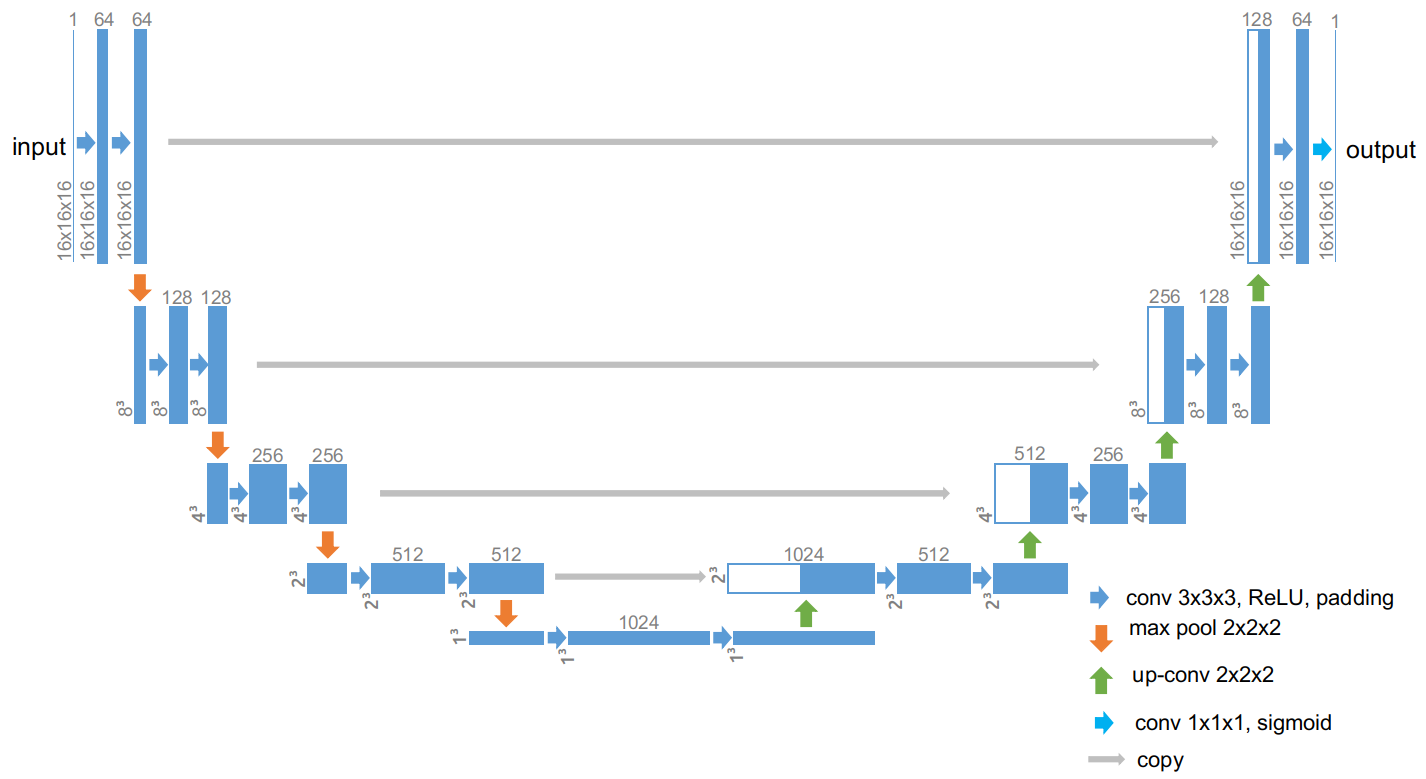} % don't need the ext-name
  \caption{The structure of 3D U-Net and the detailed information of the 3D U-Net we used}
  \label{fig:unet}
\end{figure*}

Compared to oringally proposed 2D U-Net, 3D U-Net uses the information
between slices to make the segmentation results of adjacent slices
more coherent and smooth, while 2D U-Net loses the information between
slices~\cite{3dunet2016}. Since our CT images are 3D images, we
directly use 3D U-Net. The structure of 3D U-Net and the detailed
information of the 3D U-Net we used are shown in
Figure~\ref{fig:unet}.

%%figure

Moreover, it is worth mentioning that many variants of U-Net have been
proposed since the original version. While improvements are made in
certain scenario, we only use the vanilla U-Net to demonstrate the
benefit of proposed MCMC guided network. The proposed framework can
certainly be synergized with more specifically designed U-Net or other
CNN structures.

%%%%%%%%%%%%%%%%%%%%%%%%%%%%%%%%%%%%%%%%%%%%%%%%%%%%%%%%%%%%%%%%%%%%%%
\subsection{Prior guided segmentation}\label{sec:method:seg}
Given a new image $I$ to be segmented, it is first registered, with an
optimal affine transformation $\hat{T}$, to the $\tilde{I}$ in the training
image, i.e.,
\begin{align}
  \tilde{I} \approx I\circ \hat{T} =: \hat{I}
\end{align}

Once registered, the image $\hat{I}$ falls into the same domain as
the prior map $P(\bx)$. As a result, similar MCMC sampling scheme can
be used to generate sample regions from the testing image. Denote the
regions to be $\bm{r}_i\subset \bR^3, i=1, \dots, R$.

Applying the trained CNN model to the image patches
$U_i:=\hat{I}|_{\bm{r}_i}$, we get the segmentation $V_i$'s. Note that
each $V_i$ only fills a portion of the entire image domain. In order
to form a segmentation $W(\bx)$ for the entire domain, we use a voting
scheme:
\begin{align}
  W(\bx)|_{\bm{r}_i} = W(\bx)|_{\bm{r}_i} + V_i
\end{align}

However, it is noted that the $\bm{r}_i$ are not uniformly sampled
from the entire domain, and different patches do overlap. As a result,
certain regions in the domain may have a higher voting simply due to
more patches are taken from that location. To mitigate this bias
effect, a ``sample prior'' $K$ is constructed as:
\begin{align}
  K(\bx)|_{\bm{r}_i} = K(\bx)|_{\bm{r}_i} + \bm{1}(\bx)|_{\bm{r}_i}
\end{align}

The final segmentation for $\hat{I}$ is therefore
\begin{align}
  \hat{J}:=W/K
\end{align}
and that for $I$ is
\begin{align}
  J = \hat{J}\circ \hat{T}^{-1}
\end{align}

%%%%%%%%%%%%%%%%%%%%%%%%%%%%%%%%%%%%%%%%%%%%%%%%%%%%%%%%%%%%%%%%%%%%%%%%%%%%%%%%
\section{Implementation, Experiments and Results}\label{sec:some}
We detail the algorithm implementation and experiments in
Sections~\ref{sec:some:Experiments},
and~\ref{sec:some:Implementation}, respectively.

%%%%%%%%%%%%%%%%%%%%%%%%%%%%%%%%%%%%%%%%%%%%%%%%%%%%%%%%%%%%%%%%%%%%%%
\subsection{Implementation}\label{sec:some:Experiments}
In Section~\ref{sec:some:Experiments:data}, we detail the necessary
data preprocessing, including the image registration. After that, we
generate the prior distribution, train the 3D CNN, and perform
segmentation in Section~\ref{sec:some:Experiments:final}.

%% we use 3D U-Net to classify
%% pixels in this candidate region firstly, then we transform this output
%% of the 3D U-Net to the final segmentation and this step's principle
%% are presented in Section~\ref{sec:method:seg}. The structure of 3D
%% U-Net are described in Section~\ref{sec:method:train}. The
%% Section~\ref{sec:some:Experiments:region} and
%% Section~\ref{sec:some:Experiments:final} detail the contents of
%% Section~\ref{sec:method:train} and Section~\ref{sec:method:seg}.

%%%%%%%%%%%%%%%%%%%%%%%%%%%%%%%%%%%%%%%%
\subsubsection{Data Preprocessing}\label{sec:some:Experiments:data}
In this work, the image registration is performed using the DeedsBCV
library~\cite{deformableRegistration2017}. As discussed in
Section~\ref{sec:method:learn}, the affine registration of DeedsBCV is
used, which takes about one minutes for a 3D registration task. The
registered training mask images forms the prior $P(\bx)$ in
Eq~\ref{eq:spatialPrior}.

%% When we use deedsBCV to
%% registrate images, all input images are required to be in nifti format
%% and have the same dimensions.  In addition, the method does not take
%% anisotropic voxel spacing into account, so we resample all images into
%% isotropic voxel spacing. Then, all images are cropped onto the same
%% dimensions. All preprocessings are done by convNet3D.

%% %%%%%%%%%%%%%%%%%%%%%%%%%%%%%%%%%%%%%%%%%%%%%%%%%%%%%%%%%%%%%%%%%%%%%%
%% \subsubsection{Image Registration}\label{sec:some:Experiments:registration}
%% DeedsBCV is a registration tool that is based on markov
%% random field~\cite{deformableRegistration2017,multiModel2015}.
%% And for our images, each registration process takes
%% only one minute on deedsBCV. DeedsBCV has four
%% binaries: deedsBCV(nonlinear registration), linearBCV(affine pre-alignment),
%% applyBCV(transform another short integer segmentation) and
%% applyBCVfloat(transform another grayvalue scan).
%% In our work, we use linearBCV and deedsBCV only.

%% First, we can get the affine matrix, which initialises the deformable
%% registration by applyBCV. Then, deedsBCV makes nonlinear
%% registration realized. After these steps, the moving image
%% are registrated to fixed image, the generated image is $M$.
%% And the moving image's label of pancreas are registrated to
%% fixed image, the generated label image is $L$.
%% $L$ is the pancreas segmentation of fixed image, but the
%% accuracy of this segmentation is unsatisfactory.

The shape of pancreas varies greatly in some people. So that the
accuracy of some segmentatioan is not satisfactory by using single
moving image to registrate. In order to improve the accuracy, we try
to use multiple moving images registrate to fixed image. Adopting the
multi-atlas idea, We use $N$ moving images to register to one fixed
image and botain $N$ results of pancreas segmentation about this fixed
image. The resulting prior image $P$ can be thresholded to form a
banary image with a threshold~$d$:
\begin{align}
  R(x)=\begin {cases}
    0 ,\quad P(x)<d \\
    1,\quad P(x)\ge d
  \end{cases}
\end{align}
$R$ is the new image which is a banary image.But $R$ is just a rough segmentation of pancreas,too.

%% %%%%%%%%%%%%%%%%%%%%%%%%%%%%%%%%%%%%%%%%%%%%%%%%%%%%%%%%%%%%%%%%%%%%%%
%% \subsubsection{Candidate Region Generation}\label{sec:some:Experiments:region}
The false negative, false positive, true negative and true positive
should be classfied by 3D U-Net. But pancreas is surrounded by many
organs,this leads to the peripancreatic morphology being variable, so
that this classification work is more difficult. What's more, pancreas
is small in abdomen, so true negatives are far more than the others, and
the most of these true negatives have nothing to do with the features
of pancreas and the border of pancreas. This lead the input of network
does not adequately represent the morphological features of the pancreas.

In order to reduce the burden of network's classification work and
make sure cover all pancreas region, we expand the pancreatic region
in the image $R$ by $k$ pixels. After this step, there are false
positives, true positives and true negatives in the new image $E$.
The region of pancreas in $E$ is the region where we extract patches,
so that the input of network is just the morphological feature of
pancreas and the edge feature of pancreas. Besides, pancreas is small
in the abdomen so that the number of true positives are much more than
others, this candidate region could avoid this problem. Patches are
extracted on the corresponding raw image and ground truth.

%%%%%%%%%%%%%%%%%%%%%%%%%%%%%%%%%%%%%%%%%%%%%%%%%%%%%%%%%%%%%%%%%%%%%%
\subsubsection{3D U-Net Output to Final Segmentation}\label{sec:some:Experiments:final}
\begin{figure}[htb]
  \centering
  \includegraphics[width=3.2in]{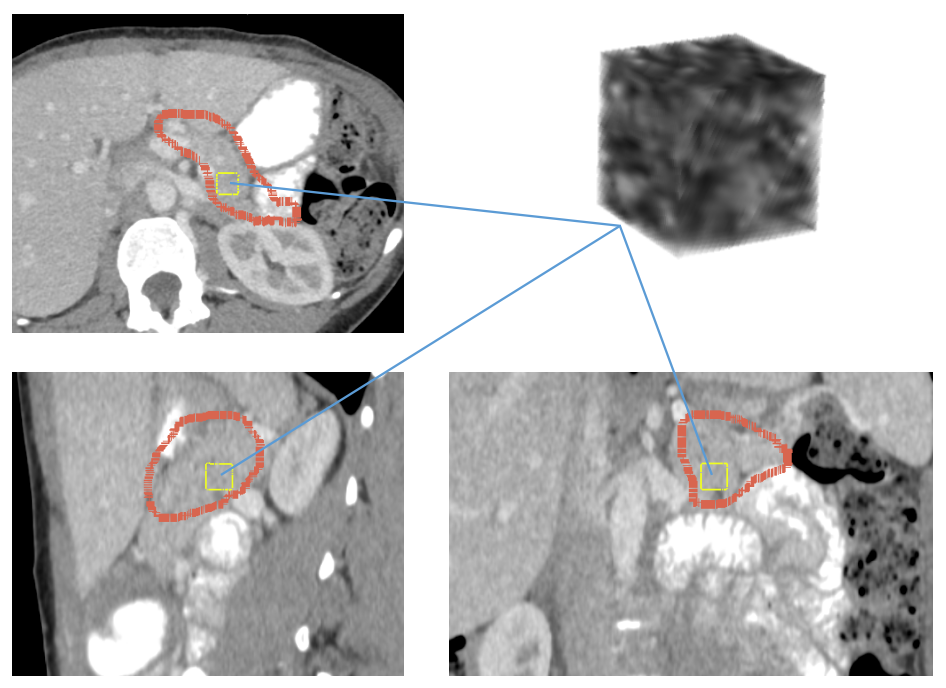} % don't need the ext-name
  %% \caption{Pick up patch in the candidate region, the candidate region
  %%   is marked as red curve. The patch region is marked as yellow curve.}
  \caption{3D patches drawn from the image}
  \label{fig:patch}
\end{figure}

Once the prior is formed, the MCMC samples drawn from $P(\bx)$ guide
the CNN training. In particular, the U-Net has the structed given in
Figure~\ref{fig:unet}. The 3D pathes extracted from the image have a
size of $16\times{}16\times{}16$, depicted in Figure~\ref{fig:patch}.

%% In our work, we traverse every image 20 times, and we pick up
%% patch(shown in Figure~\ref{fig:patch}) randomly every time, thus we get multiple results outputed
%% by network, and every result is a segmentation of pancreas.
%% For these multiple results, we superimpose these results firstly so
%% that we get an image $M$ which is similar to probobility map. Then, we
%% set a threshold $f$ of pixel values:
%% \[ W(x)=\begin{cases}
%%     0 ,\quad M(x)<f \\
%%     1,\quad M(x)\ge f
%% \end{cases} \]
%% Finally, we get the final segmentation of pancreas $W$.

%%figure

%%%%%%%%%%%%%%%%%%%%%%%%%%%%%%%%%%%%%%%%%%%%%%%%%%%%%%%%%%%%%%%%%%%%%%
\subsubsection{Evaluation Criteria}\label{sec:some:Implementation:evaluation}
In the experiments, we evaluate the results by Precision~(positive
predictive value), Recall~(sensitivity), dice similarity coefficient
(DSC), and Jaccard similarity coefficient. $tp$, $fn$, $fp$, and $tn$
represent the number of true positives, false negatives, false
positives, and true negatives, respectively. Some commonly used values
are listed below:

Precision is the proportion of positives correctly predicted among all positives predicted in prediction image.
\begin{align}
  Precision = \frac{tp}{tp + fp}
\end{align}
Recall is the proportion of positives correctly predicted among all positives in ground truth.
\begin{align}
  Recall = \frac{tp}{tp + fn}
\end{align}
Dice similarity coefficient(DSC) is a statistic used to measure the similarity of prediction image and ground truth.
\begin{align}
  DSC = \frac{2\times tp}{(tp+fp)+(tp+fn)}
\end{align}
Jaccard similarity coefficient is a statistic used to measure the similarity and diversity of prediction image and ground truth.
\begin{align}
  Jaccard = \frac{tp}{(tp+fp)+(tp+fn)-tp}
\end{align}

%%%%%%%%%%%%%%%%%%%%%%%%%%%%%%%%%%%%%%%%%%%%%%%%%%%%%%%%%%%%%%%%%%%%%%
\subsection{Experiments}\label{sec:some:Implementation}
\subsubsection{Dataset and Pre-processing}\label{sec:some:Implementation:datapre}
To facilitate the comparison of results across different publications,
we use the dataset provided by
NIH~\cite{roth2016data,deeporgan2015,clark2013cancer}. The dataset
contains 82 abdominal contrast enhanced 3D CT images and has been
manually labeled the segmentations of pancreas as ground-truth
slice-by-slice. Among them, 72 are picked for training and the
remaining 10 are used testing.

%% When we registrate images by deedsBCV, it requires that the images are
%% anisotropic in space, and the images' dimension must be consistent.
%% Therefore, all 82 images must be resampled and cropped first.
%% First, we resample all images.
%% After resampling, the size of pixels are $0.7mm\times 0.7mm\times 0.7mm$.
%% Then, we cropped all images.
%% The dimension of all images are cropped into $300\times 250\times 190$.

%%%%%%%%%%%%%%%%%%%%%%%%%%%%%%%%%%%%%%%%%%%%%%%%%%%%%%%%%%%%%%%%%%%%%%
\subsubsection{Parameter optimization}\label{sec:some:Implementation:details}
\begin{figure}[htb]
  \centering
  \includegraphics[width=3.2in]{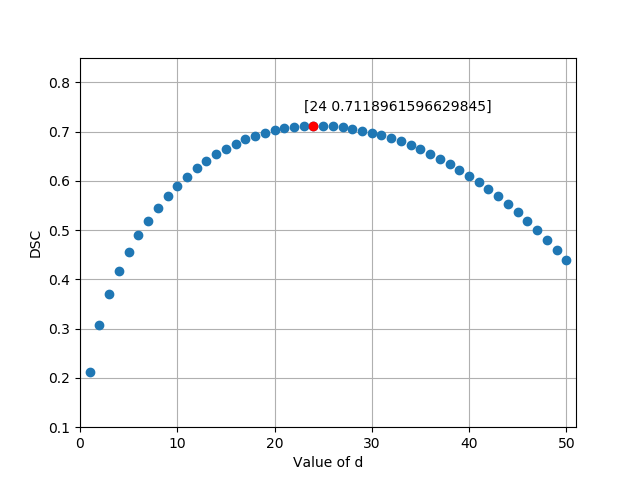} % don't need the ext-name
  \caption{the average DSC changes with different values of $d$}
  \label{fig:registrationVote}
\end{figure}

After the images registration, we get the prior distribution $P$. In
order to get banary image $R$, we set a threshold $d$ to classify
pixel value as 0 and 1. To determine the value of $d$, an image is
randomly selected and its pixels are classified into 0 and 1 by $d$
whose valuse is in (0, 72). We set $d$'s value in (1, 50), then
compute the average DSC of different values of $d$. The result is
shown in Figure~\ref{fig:registrationVote}. It is found that when the
value of $d$ is 24, the DSC between $R$ and ground truth is maximum,
so we set threshold $d$ as 24.

%%figure

In the next step, we expand the pancreatic region in the image $R$ by
$k$ pixels. It is found that when the value of $k$ is 5, the candidate
region contains most of the pancreatic region. In this candidate
region, the non-pancreatic region is in a suitable range. So we set
$k$ as 5.

%%figure
\begin{figure}[htb]
  \centering
  \includegraphics[width=3.2in]{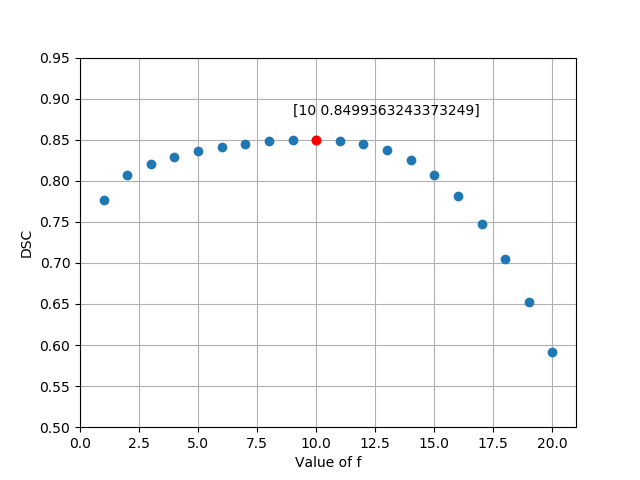} % don't need the ext-name
  \caption{The average DSC changes with different values of $f$}
  \label{fig:unetVote}
\end{figure}

As for the 3D U-Net, we set the patch size as $16\times 16\times 16$,
and set batch size as 100.  In our experiment, we use the binary cross
entropy as loss function.

Finally, we search the threshold $f$ in $\{1, \dots, 20\}$ for the
best average DSC. The result is shown in Figure~\ref{fig:unetVote}. As
can be seen, the average DSC of train data is the highest when $f=10$.

%%%%%%%%%%%%%%%%%%%%%%%%%%%%%%%%%%%%%%%%%%%%%%%%%%%%%%%%%%%%%%%%%%%%%%%%%%%%%%%%
\subsection{Experimental Results}\label{sec:some:results}
In this work, we proposed a pancreas segmentation method based on MCMC
guided deep learning. These method effectively reduces the burden of
network training, as well as allowing the network to locate the target
more robustly. Benefitting from these, our model finally get a
competitive result of an average 82.65\% Recall value and an average
78.13\% DSC value. The training of the 3D-UNet takes 8 hours for 50000
epochs on a GPU(Nvidia GTX Titan X).

%%tabel performance after sec registration vote
\begin{table}[htb]
    \centering
    \begin{tabular}{llll}
        \hline
        & Precision(\%)  & Recall(\%)  & DSC(\%) \\
        \hline
       Mean      & 61.64          & 74.48       & 66.54 \\
        Min       & 38.11          & 40.42       & 44.82 \\
        Max       & 80.58          & 88.45       & 80.25 \\
        \hline
    \end{tabular}
    \caption{Performance after multi-atlas in testing images}
    \label{tab:registration}
\end{table}

Table~\ref{tab:registration} shows the performance after
multi-atlas. The mean performance is 74.48\% Recall after the 
process of multi-atlas which means that the position of pancreas 
is located roughly in abdomen.

%%table performance after expand
%%\begin{table}[htb]
%%    \centering
%%    \begin{tabular}{llll}
%%        \hline
%%        & Precision(\%)  & Recall(\%)  & DSC(\%) \\
%%        \hline
%%        Mean      & 32.13          & 93.04       & 47.34 \\
%%        Min       & 19.15          & 72.24       & 31.97 \\
%%        Max       & 43.75          & 99.58       & 59.61 \\
%%        \hline
%%    \end{tabular}
%%    \caption{Performance of candidate region}}
%%    \label{tab:expand}
%%\end{table}

Besides, as for the extracted candidate regions, there are 90\% 
of cases being above 88.94\% Recall with the mean Recall reaching 93.04\% Recall. 
Based on this, it can be seen that most of the pancreas region is covered in
these candidate regions.

%%table performance training
\begin{table}[htb]
    \centering
    \begin{tabular}{llll}
        \hline
        & Precision(\%)  & Recall(\%)  & DSC(\%) \\
        \hline
        Mean      & 80.20          & 91.13       & 84.99 \\
        Min       & 70.56          & 60.67       & 69.35 \\
        Max       & 89.07          & 98.57       & 91.47 \\
        \hline
    \end{tabular}
    \caption{Model's performance in training images}
    \label{tab:train}
\end{table}

%%table performance test
\begin{table}[htb]
    \centering
    \begin{tabular}{llll}
        \hline
        & Precision(\%)  & Recall(\%)  & DSC(\%) \\
        \hline
        Mean      & 74.64          & 82.65       & 78.13 \\
        Min       & 56.15          & 65.99       & 66.50 \\
        Max       & 84.12          & 93.81       & 87.49 \\
        \hline
    \end{tabular}
    \caption{Model's performance in testing images}
    \label{tab:test}
\end{table}

Table~\ref{tab:train} shows the model's performance in training
images, and Table~\ref{tab:test} shows the model's performance in
testing images. Moreover, only one outlier cases has a Recall value 
below 70\%. For the Precision measurement, there are 80\% of
cases being above 71.56\% Precision, respectively. It can be seen 
that the Recall of our model is higher than precision. It may indicate 
that our model could reserve pancreas area excellently in the 
prediction processes.

%%figure
\begin{figure}[htb]
  \centering
  \includegraphics[width=3.5in]{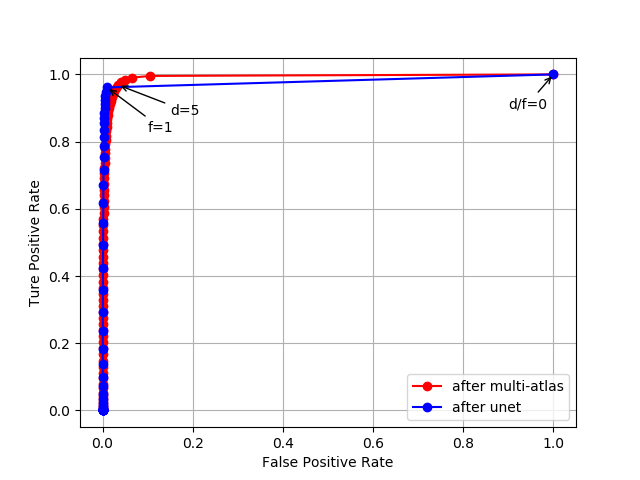} % don't need the ext-name
  \caption{The red curve shows the ROC curve after the process of 
    multi-atlas, and the blue curve is the ROC curve which is 
    from the process of U-Net, the red curve is the ROC curve 
    which is from the process of multi-atlas}
  \label{fig:roc}
\end{figure}
%%figure
\begin{figure}[htb]
  \centering
  \includegraphics[width=3.5in]{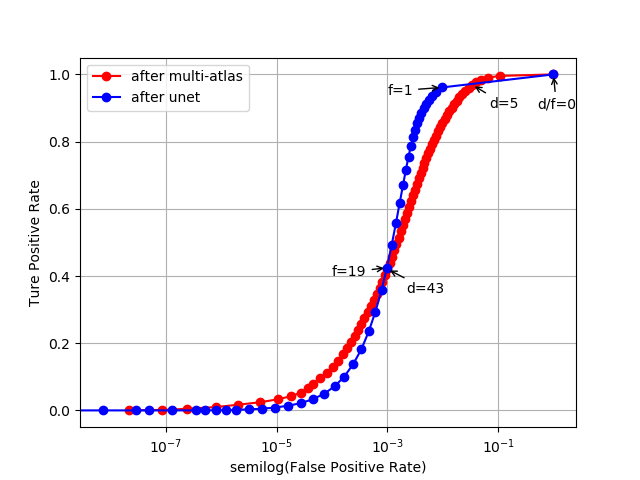} % don't need the ext-name
  \caption{ROC curves which make false positive rate be semilog.}
  \label{fig:zoomroc}
\end{figure}

The average Hausdorff distance of our predict cases and ground truths
is 11.90mm. And the maximum Hausdorff distance of our predict cases is
23.88mm, the minimum Hausdorff distance of our predict cases is
4.49mm.

%%figure
\begin{figure*}[htb]
  \centering
  \includegraphics[width=6.4in]{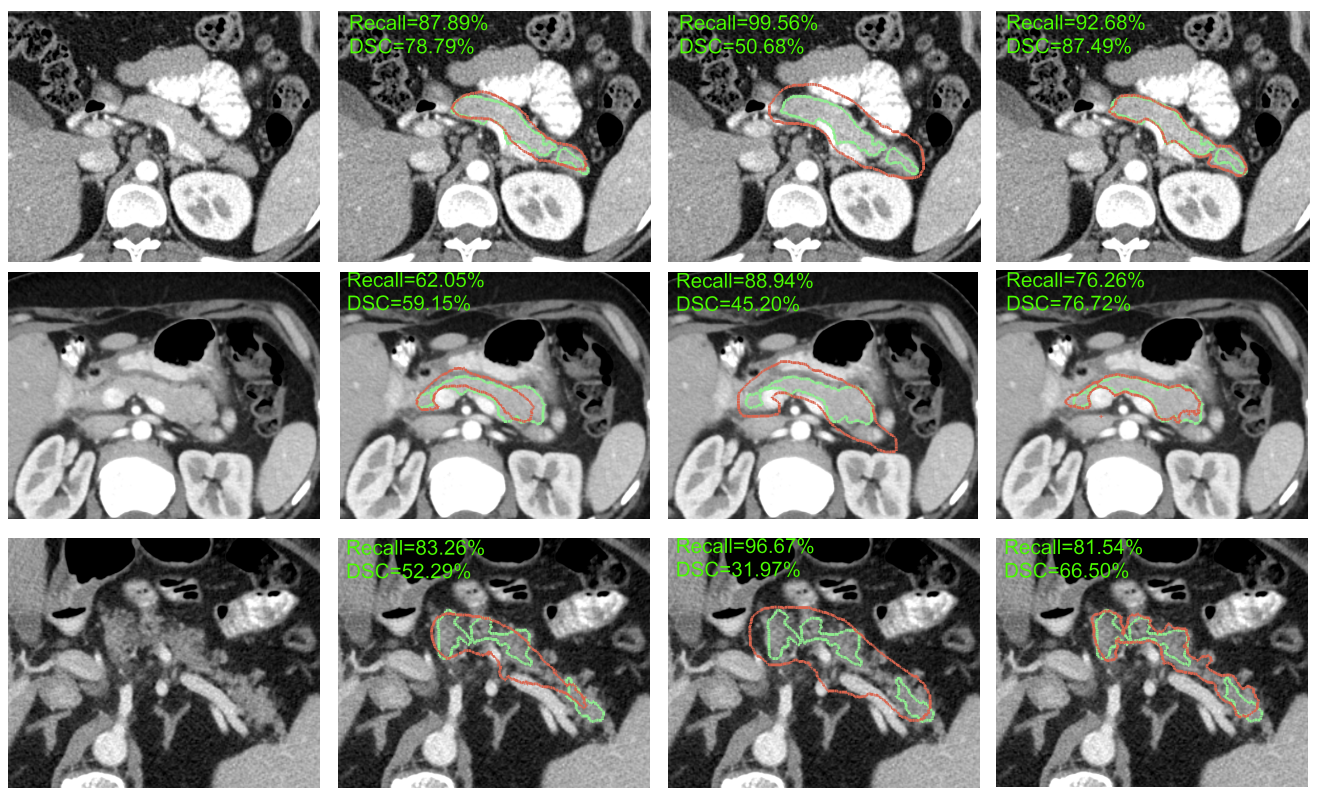} % don't need the ext-name
  \caption{Displays three examples of the outputs from
    the process of multi-atlas, the candidate regions and
    the final segmentation results. And we compare these
    outputs to ground truth, the ground truth is marked as green
    curve, and the output is marked as red curve. From left to right,
    these images are the original CT images, the outputs from
    the process of multi-altas, the candidate regions and the final 
    segmentation results}
  \label{fig:result}
\end{figure*}
%%figure
\begin{figure*}[htb]
  \centering
  \includegraphics[width=6.4in]{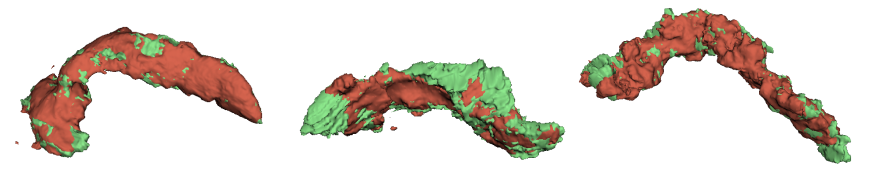} % don't need the ext-name
  \caption{Displays three examples' results in 3D rendering, this three 
    examples have been shown in Figure~\ref{fig:result}, the ground 
    truth is marked as green volume, and the output is marked as red volume.}
  \label{fig:3dresult}
\end{figure*}

Figure~\ref{fig:roc} shows the ROC curves after the process of 
multi-atlas and the process of U-Net. But the left side of this 
picture is too crowded to see clearly, so we make false positive 
rate be semilog. In Figure~\ref{fig:zoomroc},we find that 
when $0<f<19$, the ROC curve which is from the process of U-Net 
is above the ROC curve which is from the process of multi-atlas 
distinctly. These reveal that when we extract the candidate regions 
by $f=10$, our model successfully reject substantial amount of 
false-positive regions.

%%表格 5
\begin{table*}[htb]
    \centering
    \begin{tabular}{lllll}
        \hline
        & \multicolumn{4}{c}{Mean[min, max]} \\
        \cmidrule{3-5}
        Method   &Cases  & Precision(\%)  & Recall(\%)  & DSC(\%) \\
        \hline
        Roth et al., MICCAI'2015~\cite{deeporgan2015}      & 82    & -                    & -                   & 71.8[25.0$\sim$86.9] \\
        Roth et al., MICCAI'2016~\cite{deeporgan2015}      & 82    & -                    & -                   & 78.01[34.11$\sim$88.65] \\
        Zhou et al., MICCAI'2017~\cite{fixedPoint2017}     & 82    & -                    & -                   & {\bf 82.37}[62.43$\sim$90.85] \\
        Karasawa et al.,SCI'2017~\cite{multiAtlas2017}     & 150   & -                    & -                   & 78.5 \\
        Farag et al.,IEEE'2017~\cite{bottomUp2017}         & 80    & 71.6[34.8$\sim$85.8]     & 74.4[15.0$\sim$90.9]     & 70.7[24.4$\sim$85.3] \\
        Ours                                               & 82    & {\bf 74.64}[56.15$\sim$84.12]          & {\bf 82.65}[65.99$\sim$93.81]  & 78.13[66.50$\sim$87.49] \\
        \hline
    \end{tabular}
    \caption{Some recent state-of-the-art methods on pancreas
      segmentation are compared}
    \label{tab:compare}
\end{table*}

In Figure~\ref{fig:result}, we shows three examples of outputs from
the outputs from the process of multi-atlas, the candidate regions and
the final segmentation results in testing. The first row is
the best performance image with 87.49\% DSC, 92.68\% Recall and 82.85\%
Precision. The second row shows the image whose performance is close to 
the everage performance with 76.72\% DSC, 76.26\% Recall and 77.19\% 
Precision. The third row is the worst performance image with 66.50\% 
DSC, 81.54\% Recall and 56.15\% Precision. And in Figure~\ref{fig:3dresult}, 
we show the 3d rendering of this three examples. From right ro left, 
the DSC of these examples are increasing. As can be seen, we can
approximately locate the pancreas' region in its vicinity after
the process of multi-atlas, but there are some pancreas regions that are 
missed. With the process of extracting candidate regions, more pancreas regions is
included correctly. In addition, the marginal region of pancreas is
included so that we could get the marginal information of pancreas.

The results are compared with some recent state-of-the-art methods on
pancreas segmentation. Comparing the DSC in diffrernt models, our
model's lowest DSC is higher than other models substantially. What's more, 
our model's lowest Reacll and Precision are higher significantly with the 
mean Recall and Precision being similar. These show that our model is robust.
 
%%%%%%%%%%%%%%%%%%%%%%%%%%%%%%%%%%%%%%%%%%%%%%%%%%%%%%%%%%%%%%%%%%%%%%%%%%%%%%%%
\section{Conclusion, Discussion, and Future Directions}\label{sec:discussion}
In this work, we proposed a general purpose segmentation framework
that uses the Monte Carlo Markov Chain~(MCMC) to guide segmetnation of
the 3D images. Specifically, the prior spatial distribution is learned
and an MCMC scheme is utilized to generate samples from the
prior. Such samples are used to guide the sampling of patches from the
training images, which are input to the convolutional neural
network. During the segmentation, the MCMC is employed again to sample
from the high probability regions in the target image. The sampled
regions are fed to the trained CNN, from which the final segmentation
concensus are constructed. The proposed framework is applied to the
abdonimal CT images to extract the pancreas, and an average Recall
value of 82.65\% with and an average DSC value of 78.13\% are achieved.

Future directions include investigating the variances induced by the
imaging parameters, such as the field of view, with/without contrast
agent, slice/slab thickness, etc. Moreover, the proposed method will
be used in conjunction with the classifcation of various pancreatic
diseases.

%%%%%%%%%%%%%%%%%%%%%%%%%%%%%%%%%%%%%%%%%%%%%%%%%%%%%%%%%%%%%%%%%%%%%%%%%%%%%%%%
\section*{Acknowledgment}
YG would like to thank the support from the National Natural Science
Foundation of China No. 61601302 and Shenzhen peacock plan
(NO.KQTD2016053112051497). XY would like to thank the support from
the Faculty Development Grant of Shenzhen University No.~2018009.

%%Harvard
\small
\bibliographystyle{model2-names.bst}\biboptions{authoryear}
\bibliography{mcmcCnn}

\end{document}

%% file: preamble_gth818n.tex
\usepackage{amsfonts}
\usepackage{mathrsfs} %mathscr{} needs this
\usepackage{amsmath}
\usepackage{amssymb} %% \because \therefore need this package
\usepackage{amsthm} % for \newtheorem thus theorem, lemma, conj, etc. environments. and for proof environment.
\usepackage{graphicx}
\usepackage[final]{fixme}

\usepackage{url}
\usepackage[usenames,dvipsnames]{color}
\usepackage{algorithm,algorithmic}

\usepackage[normalem]{ulem}

\newcommand{\bm}[1]{\boldsymbol{#1}}
\newcommand{\bx}{\boldsymbol{x}} % better not do \bm{x} which depends on the previous def
\newcommand{\by}{\boldsymbol{y}} % better not do \bm{x} which depends on the previous def

\newcommand{\bR}{\mathbb{R}}
\newcommand{\nn}{\nonumber \\}

\DeclareMathOperator*{\argmin}{arg\,min}